*Research Article*

# A Web-Based Tool for Automatic Data Collection, Curation, and Visualization of Complex Healthcare Survey Studies including Social Network Analysis

José Alberto Benítez,[1] José Emilio Labra,[2] Enedina Quiroga,[3] Vicente Martín,[4] Isaías García,[1] Pilar Marqués-Sánchez,[3] and Carmen Benavides[1]

[1]*Department of Electrical and Systems Engineering, Universidad de León, Campus de Vegazana, s/n, 24071 León, Spain*
[2]*Department of Computer Science, Universidad de Oviedo, C/Calvo Sotelo, s/n, 33007 Oviedo, Spain*
[3]*SALBIS Research Group, Facultad de Ciencias de la Salud, Campus de Ponferrada, Avda Astorga, s/n, Ponferrada, 24402 León, Spain*
[4]*GIGAS Research Group, Facultad de Ciencias de la Salud, Campus de Vegazana, s/n, 24071 León, Spain*

Correspondence should be addressed to José Alberto Benítez; jbena@unileon.es

There is a great concern nowadays regarding alcohol consumption and drug abuse, especially in young people. Analyzing the social environment where these adolescents are immersed, as well as a series of measures determining the alcohol abuse risk or personal situation and perception using a number of questionnaires like AUDIT, FAS, KIDSCREEN, and others, it is possible to gain insight into the current situation of a given individual regarding his/her consumption behavior. But this analysis, in order to be achieved, requires the use of tools that can ease the process of questionnaire creation, data gathering, curation and representation, and later analysis and visualization to the user. This research presents the design and construction of a web-based platform able to facilitate each of the mentioned processes by integrating the different phases into an intuitive system with a graphical user interface that hides the complexity underlying each of the questionnaires and techniques used and presenting the results in a flexible and visual way, avoiding any manual handling of data during the process. Advantages of this approach are shown and compared to the previous situation where some of the tasks were accomplished by time consuming and error prone manipulations of data.

## 1. Introduction

Computing and Information Science play a more and more important role in healthcare studies and applications [1]. A lot of healthcare studies dealing with life habits consist in cross-sectional or epidemiological surveys using a great amount of information gathered from the subjects by means of different questionnaires. While information obtained from these studies is very useful for both research and healthcare professionals, the management and analysis of data is many times cumbersome and leads to long and tedious processes carried out by humans, which also implies the possibility of being error prone.

Data gathering is usually carried out with online surveys designed and created with well-known free (e.g., Google Docs [2], SurveyMonkey [3]) or commercial solutions. These tools provide an easy way of creating and obtaining data from questionnaires, but with a number of drawbacks. First, they are generic tools, not aimed at any specific domain, though some of them include healthcare oriented templates and provide data protection and security services. Many of them provide simple capabilities for the type of questions to be created and, moreover, they require building each question in the survey from scratch, without the possibility of reusing other questionnaires or parts of them. When obtaining the results, the great majority of tools store data from individual's answers in the form of plain datasheets making the later processing and visualization of data difficult, especially when dealing with complex questions. There is no tool incorporating an integrated set of features as the one developed during this research, which includes functionalities for the creation of surveys, the processing of data, and the visualization and study of the results.

Data curation, once it has been obtained from the surveys, is usually left to the professional/researcher that designed the questionnaire. They are responsible for obtaining the information in the format that is useful for their work. This process is most of the times a hand-made one and so it is usually very time consuming and error prone.

There are a number of specialized questionnaires that could be reused without the need of creating them from scratch. These questionnaires are used for obtaining a number of psychosocial measures regarding individuals. Some examples are FAS II (Family Affluence Scale) [4], AUDIT (Alcohol Use Disorders Identification Test) [5–7], KIDSCREEN 27 (for studying quality of life perception among young people) [8, 9], or ESTUDES (a national, Spanish, survey including substance consumption indicators and life habits) [10]. All of these questionnaires have a well-known set of questions and a fixed scoring and interpretation based on numerical scales and quantitative-to-qualitative mappings when total scores are calculated. So, both the integration of these questionnaires into custom, broader ones, and the curation of data obtained from the corresponding responses can be automated, facilitating the work of the users and eliminating errors due to human factors.

When including social network analysis (SNA) in healthcare studies the former considerations acquire greater relevance. Obtaining social relationships from surveys require the use of complex questions where the participants usually are arranged into matrix-like structures. Results from these kinds of questions consist in a great amount of interrelated data that is quite difficult to handle manually, especially when dealing with hundreds or thousands of individuals. Moreover, when the same questionnaire is used by different groups of people the results have a different number of rows and, which is more problematic, columns, in the resulting spreadsheet, which makes it even more difficult to handle. The proper representation of this kind of information is crucial for the later phase of analysis and visualization because algorithms used for social network analysis must be run by computers due to their hardness and execution time.

Another level of complexity regarding the gathering and processing of data appears if the study of a given set of people is carried out at different points in time, that is, if it is desired to know how the social relationships and metrics evolve in time. These kinds of studies are popular in today's approaches to social-based substance consumption research.

This research aims to study, design, and develop a solution able to automate every step in the process of questionnaire design and deployment, data gathering and curation, and later analysis and visualization, in order to reduce the drawbacks and difficulties of the manual handling that have been previously exposed and help professionals/researchers to focus on their work and not on time consuming, error prone tasks.

In order to achieve these objectives, a number of techniques and tools have been used and created, allowing the capture and representation of data into the computer and showing it in a friendly graphical user interface. Any healthcare professional or researcher, without deep knowledge of computing, is able to design and build a personal questionnaire, publishing it in order to gather data and see the results in the more convenient way for the purpose it has been constructed.

The ideas and tools described in this research have been applied to a study about influence mechanism on alcohol consumption among adolescents that uses a questionnaire for gathering data about the student psychosocial situation, his/her life perception and socioeconomical position, alcohol consumption habits, and friendship and family network. This study uses a complex questionnaire and techniques from social network analysis (SNA) in order to find and demonstrate the mentioned influence mechanisms, as well as for obtaining a picture of the situation of each student and his/her class regarding alcohol consumption levels.

The rest of the paper is organized as follows. Section 2 is devoted to giving an overview of the application that has been built, introducing some of the key features that help to automate and improve the manual processes that usually take place when performing a healthcare study. Section 3 presents some discussion about the results that have been obtained, showing the advantages of this approach over the manual method. Finally, Section 4 shows the conclusions of this research, showing how collaborative work of healthcare and computer science researchers may help to build useful applications for e-health.

## 2. Materials and Methods

*2.1. Standardized Forms.* Current research on substance consumption and abuse tries to study a wider range of factors that may be involved in consumption habits. In particular, the social environment where the individual is immersed is of special interest. This is even more important when the population subjects of study are adolescent people because it is a stage in life where close friends and colleagues may influence each other's life style and habits, including alcohol and substance consumption (a survey on this kind of studies for the case of the European Union can be found in [11]). An approach based on the analysis of traditional data about individual consumption levels and risk assessment along with techniques from social network analysis applied to friendship and family relationships may give more insight into the individual and group situation than narrower traditional studies.

Traditional tools to gather information about individual consumption habits and other psychosocial and socioeconomic measures for an individual include a number of well-known, validated, and standardized questionnaires. Some of them have been included in the solution proposed in this research:

(i) AUDIT (Alcohol Use Disorders Identification Test) [12]: this test is used to detect problematic levels of alcohol consumption or dependence.

(ii) FAS II (Family Affluence Scale II) [4] is used for assessing family wealth.

(iii) ESTUDES (Poll about Drug Use in Secondary School in Spain) [10]: it is a biannual study for gaining insight

Table 1: AUDIT risk level scoring [15].

| Risk level | Intervention | AUDIT score |
| --- | --- | --- |
| Zone I | Alcohol education | 0–7 |
| Zone II | Simple advice | 8–15 |
| Zone III | Simple advice plus brief counseling and continued monitoring | 16–19 |
| Zone IV | Referral to specialist for diagnostic evaluation and treatment | 20–40 |

into behaviors and attitudes about substance use. It is a test that includes questions about consumption of different substances. Those related to alcohol were discarded because they were obtained in other parts of the main questionnaire.

(iv) KIDSCREEN-27 (Health Related Quality of Life Questionnaire for Children and Young People and Their Parents) [13]: it is used for knowing the quality of life perception of the adolescent based on five scales: physical and psychological well-being, autonomy and parent relation, peers and social support, and school environment.

(v) Self-efficacy [14] (Spanish adapted version): it assesses the belief of the adolescent about his/her own capacities to achieve different goals, specially facing stressful situations.

These questionnaires have a scoring method in order to obtain a quantitative and/or qualitative result that characterizes the individual into a number of categories that can be later used for decision making or further analysis. Table 1 shows, as an example, the scoring and categories for the AUDIT test. This test consists of ten questions whose responses are graded from 0 to 4. Adding the grade corresponding to each response gives an AUDIT score (quantitative value), which is mapped to a risk level (qualitative value) and a possible intervention to be achieved. Handling these calculations manually is time consuming and may lead to errors. In other tests, it is even more complex to obtain a score, as is the case of KIDSCREEN.

### 2.2. Data and Analysis Management Application

#### 2.2.1. Introduction of Application.
The proposed application has a login system where the professional or researcher (once she has registered into the application) has access to a personalized control panel where she can perform all the tasks needed for her research. The application has been developed with a web architecture, using a web browser for the graphical user interface and the last standards for development of these kinds of applications (HTML5, jQuery, AJAX, CSS3). On the server side, it has been developed with PHP and MySQL as the database management system.

Some libraries and APIs have been used for displaying the data, especially when dealing with graphs that show social connections (e.g., SigmaJS: http://sigmajs.org/; see [16] for a similar approach) and the results of the social network analysis algorithms (e.g., the Louvain community detection algorithm described in [17]).

Three different user roles can be distinguished in the application:

(i) Super Administrator has full permissions to manage questionnaires, users, and respondents (excluding access to personal data as described in data protection laws).

(ii) Interviewer/pollster can create and edit questionnaires, using the validated test available in the platform or creating questions from scratch.

(iii) Respondent has access to the application only for filling the questionnaire.

A researcher or professional is able to accomplish the following tasks:

(i) Manage questionnaires (both validated and customized)

(ii) Create and edit individual questions or question groupings for the questionnaires

(iii) Manage interviewers

(iv) Analyze and visualize the data from the questionnaires that have already been filled

Next sections are devoted to describing each of these functionalities, showing how the application eases the work of the user.

#### 2.2.2. Standardized Form Management.
From this menu item, the user can see a listing of the different validated tests that are common in the field of healthcare studies (as is the case of the aforementioned AUDIT, FAS II, KIDSCREEN, etc.).

The user can browse and deeply inspect the questions and characteristic of the questionnaire, the responses available for each question along with their scoring, and the general score and its meaning for the whole questionnaire.

A user with Super Administration role is able to add new validated questionnaires. If this is the case, the reference where the complete description of the questionnaire is stated must be provided.

#### 2.2.3. Customized Questions Management.
Users with Super Administration or interviewer role can design, create, edit, or delete questions inside their questionnaires. For doing so there is an editing window where the he/she can see the questionnaires he/she has created and, within each one of them, questions can be added, deleted, or edited. But an interviewer is not able to see the questionnaires from other ones.

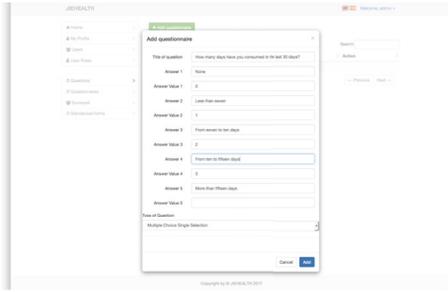

Figure 1: Adding a question.

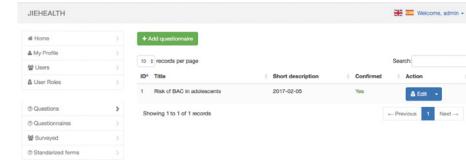

Figure 2: Section to add questionnaires.

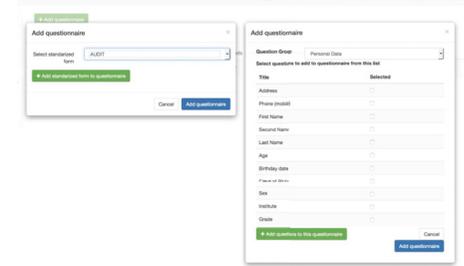

Figure 3: Modal window to add questions to questionnaire.

For the questions that have been created by the interviewer, the score can be assigned at design time. Also, questions can be grouped into sets, giving a total score for the set based on a formula including the scores of the different questions within it, in a similar way to the validated tests that exist in the platform.

The procedure for creating a new question is below (see Figure 1):

(i) Adding a question with a number of possible answers, giving a value for each one of them

(ii) Grouping a number of questions and giving a value to the total score based on a formula using the scores of the individual questions

The creator of the question can also indicate if the response given needs to be anonymized so as not to violate personal data protection laws. The system will automatically perform a substitution of data from these fields into anonymized ones.

*2.2.4. Management of Customized Questionnaires.* The users with Super Administration or interviewer role can design, create, edit, or delete questionnaires. There is a dedicated space for this, along with a listing of the questionnaires that each user is able to manage. An interviewer will only be able to manage questionnaires created by himself/herself.

Adding a questionnaire will prompt the following information (see Figure 2):

(i) The questionnaire type

(ii) A description for the questionnaire

(iii) Questions and sets of questions or validated questionnaires to be integrated in this one

(iv) Generating questions based on the group of people to be polled: this feature is especially useful for capturing social network data

When creating and editing a questionnaire, it is possible to add an existing, validated test or a group of existing questions that have already been created by someone else or to add new questions (see Figure 3).

For capturing relationships, it is possible to introduce a list of people who are going to fill in the questionnaire in order to use their data for building questions related to social relationships. As an example, when trying to find how social relationships influence alcohol consumption in adolescents, it is necessary to have a listing of students in a class in order to build questions of the type "How often do you go out for an alcoholic drink with the following colleagues."

*2.2.5. Management of Respondents.* The Super Administration or interviewer roles also allow performing a listing of people with the respondent role, with the permission for adding, editing, or deleting them. When adding new respondents, groups can also be constructed and assigned different questionnaires to be filled.

A similar questionnaire can be assigned to the same individual or group of individuals. This is used for the case where the study needs the same set of data obtained at different points in time as is the case, for example, when studying consumption habits and friendship relationships evolving in time, with the aim of finding influence or selection processes in the social environment of the individual.

*2.3. Application to Visualize Survey Data.* The application has a user interface to show the results of the different questionnaires that a user is able to manage. This part of the application shows the resulting piece of data once processed or curated, that is, once the calculations of the different scores have been performed.

As well as plain qualitative and quantitative measures obtained from data curation, there is a part of the application devoted to showing social relationships obtained with the questionnaires. This type of information is shown by means of different graphs where nodes and edges represent individuals (with sizes, shapes, and colors representing different characteristics of the individual) and relationships (friendship, drinking companions, etc.), respectively.

Traditional research using social network analysis (SNA) has been carried out by using tools like UCINET [18]. While this tool has a very complete set of analysis tools and it has been extensively used for a number of studies during the first decade of the XXI century [19], it is a tool that works in isolation in a desktop environment, with little or no possibility of being integrated into a broader system as is the case in this research. New tools exist today that have been designed with flexibility, extensibility, and integration capacities in mind, with a number of plugins or APIs that allow using them in tailored applications. In the case of this research, Gephi [20–22] has been chosen as the tool to allow the integration of social network analysis and visualization into the application (see Figure 4).

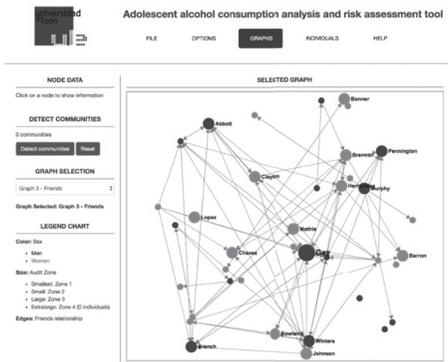

Figure 4: Application showing social network of friendship relationships.

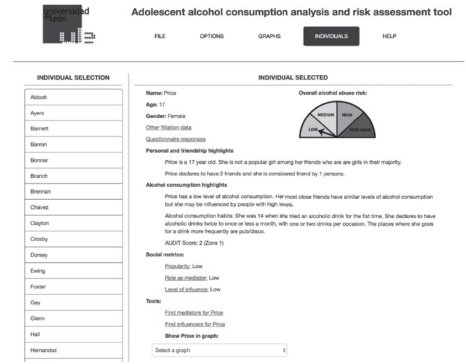

Figure 5: Application showing information about an individual.

This part of application performs the following functionalities, based on the kind of information that can be obtained from the data:

(i) Load the set of data coming from the responses to the questionnaire and show the different networks that can be of use for gaining insight into alcohol use related to friendship (at school level) and family networks.

(ii) Show the data graphs and the results from analysis regarding the social patterns regarding alcohol consumption taking into account different levels of peer relationship (acquaintance, partner, and friend) and also the family environment.

(iii) Show basic data and a report-like description of any individual in the network about his/her alcohol consumption status, specially concerning alcohol use disorder risk or any kind of relevance within the network that the individual may pose (being a mediator, an influencer for others, etc.). See Figure 5.

(iv) Show, for each individual, who can act as an influencer for him/her.

(v) Show similar report-like description stated previously but for entire social networks and groups that may be found.

(vi) Show and report the relationship that may hold between alcohol consumption, socioeconomic status, self-perception, and self-efficacy.

(vii) Show and report factors that may be related to the level of alcohol consumption, like polydrug use and consumption and relationship environments.

The application must show the information in such a format that any health professional or researcher can understand it, without need for knowing anything about social network analysis techniques or terminology. The interaction with the tool and the information presented must utilize commonly used terminology when describing adolescent characteristics, friendship or family relationships, and alcohol consumption habits.

Figure 4 shows the information displayed for the main social network of friendship relationships. In this view, circles represent individuals. Light grey colored ones are girls while dark grey ones are boys. The size of the circle represents, in this view, the AUDIT zone as resulting from the questionnaire response for each student (bigger circles mean higher levels of alcohol consumption). The user can click on any of the circles and the basic information for that student will be shown in the left-hand part of the window. One of the interesting things that can be noticed at first sight in this view is, for example, the fact that the most popular individual (the one that is named as friend by more people) is also the one with a greater AUDIT score. With the aid of different menus and items the user can easily navigate to any other graph generated during the data processing phase.

Figure 5 represents the window showing information for a given individual. This window shows the basic personal data for the individual and a graphical representation of the calculated alcohol use disorder risk level. Two natural language descriptions are also generated and presented to the user. The first one describes the particularities of the individual from the point of view of the friendship network where he/she is involved, while the second is devoted to alcohol consumption habits. Some social measures as the popularity, capacity of influence, or mediation are also presented based on the positions and connections the student has in different social networks. Finally, the user can find a number of mediators that can facilitate the information flow towards the current student or some individuals that may influence him/her. This student can also be shown in any of the different social graphs where he/she appears.

*2.4. Social Network Analysis.* One of the more difficult parts in application design and development is the part devoted to social network analysis. It was necessary to carry out a study of the different sets of techniques and algorithms used in this field in order to run the algorithms and represent the information in the more convenient way to the user.

In order to obtain relevant information from SNA techniques, the questions and responses to be included in the questionnaires must be carefully chosen [23]. Representation of social connections and information is usually achieved by matrix structures that can be square or rectangular (depending on whether they represent mode 1 or mode 2 networks). Square matrices are used for capturing relationships that are stablished on the same set of individuals, as is the case with friendship relationships within a given classroom, while rectangular ones represent relationships that are stablished between different sets of subjects, as is the case, for example, with students and the places where they use to consume alcoholic drinks.

Further distinctions may be made within each category depending on the characteristics of the relationship. For example, we can ask students to name their best friends, but this relationship may not be reciprocated by the people they name as their friend. Also, we can ask each student to weight the friendship relationship (according to a scale going from "I know him" to "he is my best friend," e.g.) [24]. All these considerations show the difficulty to build a model for the representation of the information and also for the design and construction of questionnaires.

Once relationships stablished among different sets of students are captured, a number of algorithms may be run in order to obtain interesting information about individuals or groups. Different measures can be obtained for each individual showing their relevance in the network of friends, for example, what could involve influence others regarding alcohol consumption. It is also important to detect groups of people having strong connections among them. Clustering algorithms have also been used in order to show this information [17, 25].

## 3. Results and Discussion

The application that has been presented in this paper was developed during a collaborative effort where healthcare and computer science researchers worked together with the aim of facilitating studies about alcohol consumption in adolescent population combining traditional alcohol consumption habit measures with metrics and tools from social network analysis. The objective was to cover every step of the process, from the design and creation of the questionnaires used to gather data to the presentation of the resulting sets of information, hiding the cumbersome calculations used for scoring the different tests and the complex concepts and algorithms used by SNA techniques.

This collaborative work had, as initial trigger, the difficulties found by the healthcare team when facing a study where a complex questionnaire was created from scratch, consisting of 252 questions for gathering data about alcohol consumption and friendship relationships on different secondary schools. The questionnaire was initially created by an online generic tool. It was later filled by a total of 214 students from 9 classrooms across 3 different secondary schools. A total of 145520 questions were answered and stored in a plain spreadsheet table.

Some of the difficulties that were found during the process were as follows.

(i) Existing validated tests like AUDIT, FAS II, and others had to be introduced one question at a time when creating the questionnaire. This led to mistakes or omissions that had to be detected and solved in a time consuming reviewing process; and, even after this process, some of the questions remained with a number of misspelling errors when the questionnaire was published.

(ii) Questions created for gathering a number of relationships (friendship, consumption companions, family relationships, etc.) had to be constructed by introducing the list of involved students again and again; even with using some copy-paste tricks the process was quite tedious and time consuming. Moreover, if a mistake was made in the name of a student or if one student should be removed from the questionnaire (or a new one added) once it had been already introduced, then a careful editing of all the questions had to be accomplished.

(iii) Once one questionnaire was completed, there was no way to use a part of it in another, new one; the only solution was to copy and paste the whole questionnaire and edit the copy.

(iv) Once the questionnaires were filled, researchers had to spend about 30 minutes for each of the respondents in order to obtain the scores from their responses. Moreover, data must be processed by people who knows how each of the questions or validated tests score.

(v) Social network analysis tools like Gephi [20], Pajek [26], and UCINET [18] were used (as standalone applications) for obtaining network graphs from the collected data, spending about two hours per classroom in the process. Some mistakes were also made in the process of moving data from the spreadsheet to the given tools. Moreover, only people expert in the field of social network analysis could perform this step, and the results were only meaningful to them.

(vi) The final resulting set of processed data consisted in isolated pieces of information, that is, graphs displaying social relationships and spreadsheets containing scores from the questionnaire. There was no easy way to navigate the results or to query for a given set of individuals with a given score. It was very difficult to obtain a good insight into the situation of the studied individuals and groups regarding their alcohol consumption situation and behavior.

(vii) It was clear that if the questionnaire was to be presented again to the same set of students in order

to study their evolution of consumption or friendship in time, the results from one of the experiments were going to be very difficult to compare with others.

The application developed during this research helps to solve the aforementioned problems. The development was time consuming but the advantages are clear, especially when the system can be used by many studies from now on:

(i) Validated tests are now stored in the platform, with the set of questions and their corresponding response values, along with the formulae used to obtain the final score. Other questions that were created from scratch can also be reused for future questionnaires.

(ii) Questions aimed at obtaining social relationships can now be fed with individual data by combining a listing of the involved individuals with the given question. If an individual is added, removed, or edited, the whole set of questions will be automatically updated.

(iii) Once the questionnaire is filled, scores are calculated and social network analysis techniques are applied instantaneously and with no error because the formulae and methods used are properly stored in the application.

(iv) Having integration in mind from the beginning allowed building an application where all the information is interrelated. Graphs showing social networks can be used to navigate to individual information and his/her corresponding scores and personal data just by clicking on the node icon representing the individual.

(v) Results from several applications of the same questionnaire can be easily shown in the application by means of tabs representing the different responses obtained at different times.

The application has been used for research purposes, studying a number of secondary school communities and trying to find the relationship between alcohol consumption habits and social metrics from friendship and familiar networks. The tool helped to find the main social network parameters for the different groups and individuals involved in the study, resulting in a strong connection among alcohol consumption habits and group formation (by selection processes) or leadership.

From the point of view of survey design and creation, the tool was evaluated as very useful and user friendly by the team in charge of building questionnaires, comparing their experience to the previous, manual work that they had to perform before.

## 4. Conclusions

Computing and Information Science are very important tools for today's healthcare studies and applications. The complexity of the techniques and tools used in these studies and the increase in the amount of data that can be obtained from individuals and groups make it necessary to use automated processes for the gathering, manipulation, analysis, and visualization of the information.

Much of the data that can be obtained from an individual comes from online surveys that are designed each time a study is to be accomplished. This leads to time consuming and error prone processes that can easily be automated. In this sense, collaboration among researchers of healthcare, knowledge representation, and computing science are crucial to building the tools needed to avoid this situation.

Complex tools like the one presented in this paper can help to achieve studies that can be run by healthcare professionals or researchers without the need of being experts in the techniques underlying the automated processes that the application runs internally (e.g., knowing how to score the different tests or how social network analysis is carried out).

The use of tools like the one described in this paper helps to focus on the goals of the studies and not on the data gathering or manipulation that can be easily automated. Information processing and visualization is also greatly improved if the application is properly designed to display the data in an integrated, visual, and flexible user interface.

As future line of work, the inclusion of new functionalities that could, automatically, provide insight into the situation and changes in the relationships of the same set of individuals at different points in time would be a good enhancement for the tool, as it would allow improving the usefulness of the application for research purposes. A study on how this tool may help in real scenarios is also a planned future work; the tool will be presented to a number of healthcare and education professionals in order to explore and test the possible applications and benefits of the system, obtaining valuable feedback that can be used to enrich it.

## Conflicts of Interest

The authors declare that there are no conflicts of interest regarding the publication of this paper.